# ATTac-2000: An Adaptive Autonomous Bidding Agent

**Peter Stone**                                        PSTONE@RESEARCH.ATT.COM
**Michael L. Littman**                                 MLITTMAN@RESEARCH.ATT.COM
*AT&T Labs Research, 180 Park Avenue*
*Florham Park, NJ 07932 USA*

**Satinder Singh**                              SATINDER.BAVEJA@SYNTEKCAPITAL.COM
**Michael Kearns**                              MICHAEL.KEARNS@SYNTEKCAPITAL.COM
*Syntek Capital, 423 West 55th Street*
*New York, NY 10019 USA*

## Abstract

The First Trading Agent Competition (TAC) was held from June 22nd to July 8th, 2000. TAC was designed to create a benchmark problem in the complex domain of e-marketplaces and to motivate researchers to apply unique approaches to a common task. This article describes ATTac-2000, the first-place finisher in TAC. ATTac-2000 uses a principled bidding strategy that includes several elements of *adaptivity*. In addition to the success at the competition, isolated empirical results are presented indicating the robustness and effectiveness of ATTac-2000's adaptive strategy.

## 1. Introduction

The first Trading Agent Competition (TAC) was held from June 22nd to July 8th, 2000, organized by a group of researchers and developers led by Michael Wellman of the University of Michigan and Peter Wurman of North Carolina State University (Wellman, Wurman, O'Malley, Bangera, Lin, Reeves, & Walsh, 2001). Their goals included providing a benchmark problem in the complex and rapidly advancing domain of e-marketplaces (Eisenberg, 2000) and motivating researchers to apply unique approaches to a common task. A key feature of TAC is that it required *autonomous bidding agents* to buy and sell *multiple interacting goods* in auctions of different types.

Another key feature of TAC was that participating agents competed against each other in a preliminary round and many practice games leading up to the finals. Thus, developers changed strategies in response to each others' agents in a sort of escalating arms race. Leading into the competition day, a wide variety of scenarios were possible. A successful agent needed to be able to perform well in any of these possible circumstances.

This article describes ATTac-2000, the first-place finisher in TAC. ATTac-2000 uses a principled bidding strategy, which includes several elements of *adaptivity*. In addition to the success at the competition, isolated empirical results are presented indicating the robustness and effectiveness of ATTac-2000's adaptive strategy.

The remainder of the article is organized as follows. Section 2 presents the details of the TAC domain. Section 3 introduces ATTac-2000, including the mechanisms behind its adaptivity. Section 4 describes the competition results and the results of controlled experiments testing ATTac-2000's adaptive components. Section 5 compares ATTac-2000 with some of the other TAC participants. Section 6 presents possible directions for future research and concludes.

## 2. TAC

A TAC game instance pits 8 autonomous bidding agents against one another. Each TAC agent is a simulated travel agent with 8 clients, each of whom would like to travel from TAC-town to Boston and back again during a common 5-day period. Each client is characterized by a random set of preferences for the possible arrival and departure dates; hotel rooms (The Grand Hotel and Le Fleabag Inn); and entertainment tickets (symphony, theater, and baseball). To obtain utility for a client, an agent must construct a travel package for that client by purchasing airline tickets to and from TACtown and securing hotel reservations; it is possible to obtain additional utility by providing entertainment tickets as well. A TAC agent's score in a game instance is the difference between the sum of its clients' utilities for the packages they receive and the agent's total expenditure.

TAC agents buy flights, hotel rooms and entertainment tickets in different types of *auctions*. The TAC server, running at the University of Michigan, maintains the markets and sends price *quotes* to the agents. The agents connect over the Internet and send bids to the server that update the markets accordingly and execute transactions.

Each game instance lasts 15 minutes and includes a total of 28 auctions of 3 different types.

**Flights (8 auctions):** There is a separate auction for each type of airline ticket: flights to Boston (*inflights*) on days 1–4 and flights from Boston (*outflights*) on days 2–5. There is an *unlimited* supply of airline tickets, and their *ask price* periodically increases or decreases randomly by from $0 to $10. In all cases, tickets are priced between $150 and $600. When the server receives a bid at or above the ask price, the transaction is *cleared immediately* at the ask price. *No resale* of airline tickets is allowed.

**Hotel Rooms (8):** There are two different types of hotel rooms—the Boston Grand Hotel (BGH) and Le Fleabag Inn (LFI)—each of which has 16 rooms available on days 1–4. The rooms are sold in a 16th-price *ascending* (English) auction, meaning that for each of the 8 types of hotel rooms, the 16 highest bidders get the rooms at the 16th highest price. For example, if there are 15 bids for BGH on day 2 at $300, 2 bids at $150, and any number of lower bids, the rooms are sold for $150 to the 15 high bidders plus one of the $150 bidders (earliest received bid). The ask price is the current 16th-highest bid. Thus, agents have no knowledge of, for example, the current highest bid. New bids must be higher than the current ask price. *No bid withdrawal* and *no resale* is allowed. Transactions only *clear when the auction closes*. To prevent agents from all waiting until the end of the game to bid on hotel rooms, hotel auctions can close after an unspecified period (roughly one minute) of inactivity (no new bids received).

**Entertainment Tickets (12):** Baseball, symphony, and theater tickets are each sold for days 1–4 in *continuous double auctions*. Here, agents can *buy and sell* tickets, with transactions *clearing immediately* when one agent places a buy bid at a price at least as high as another agent's sell price. Unlike the other auction types in which the goods are sold from a centralized stock, each agent starts with a random endowment of entertainment tickets. The prices sent to agents are the *bid-ask spreads*, i.e., the highest current bid price and the lowest current ask price (due to immediate clears, ask price is always greater than bid price). When a bid that beats the current bid (ask) price arrives, the sale price is the standing bid (ask) price, as opposed to the arriving ask (bid) price. In this case, *bid withdrawal* and *ticket resale* are both permitted.

In addition to unpredictable market prices, other sources of variability from game instance to game instance are the client profiles assigned to the agents and the random initial allotment of entertainment tickets. Each TAC agent has 8 clients with randomly assigned travel preferences. Clients have parameters for ideal arrival day, $IAD$ (1–4); ideal departure day, $IDD$ (2–5); grand hotel value, $GHV$ ($50–$150); and entertainment values, $EV$ ($0–$200) for each type of entertainment ticket.

The utility obtained by a client is determined by the travel package that it is given in combination with its preferences. To obtain a non-zero utility, the client must be assigned a *feasible* travel package consisting of an arrival day $AD$ with the corresponding inflight, departure day $DD$ with the corresponding outflight, and hotel rooms of the *same type* (BGH or LFI) for each day $d$ such that $AD \leq d < DD$. At most one entertainment ticket can be assigned for each day $AD \leq d < DD$, and no client can be given more than one of the same entertainment ticket type. Given a feasible package, the client's utility is defined as

$$1000 - travelPenalty + hotelBonus + funBonus$$

where

- $travelPenalty = 100(|AD - IAD| + |DD - IDD|)$

- $hotelBonus = GHV$ if the client is in the GBH, 0 otherwise.

- $funBonus =$ sum of relevant $EV$'s for each entertainment ticket type assigned to the client.

A TAC agent's final score is simply the sum of its clients' utilities minus the agent's expenditures. Throughout the game instance, it must decide what bids to place in each of the 28 auctions. At the end of the game, it must submit a final allocation of purchased goods to its clients.

The client preferences, allocations, and resulting utilities from one particular game from the TAC finals (Game 3070 on the TAC server) are shown in Tables 1 and 2.

For full details on the design and mechanisms of the TAC server, see Wellman et al. (2001).

## 3. ATTac-2000

**ATTac-2000** finished first in the Trading Agent Competition using a principled bidding strategy, which included several elements of *adaptivity*. This adaptivity gave **ATTac-2000** the flexibility to cope with a wide variety of possible scenarios at the competition. In this section, we describe **ATTac-2000**'s bidding strategy, its method for determining the best allocation of goods to clients, and its three forms of adaptivity. **ATTac-2000**'s high-level strategy is summarized in Table 3.

| Client | *IAD* | *IDD* | *GHV* | *BEV* | *SEV* | *TEV* |
|--------|-------|-------|-------|-------|-------|-------|
| 1 | Day 2 | Day 5 | 73 | 175 | 34 | 24 |
| 2 | Day 1 | Day 3 | 125 | 113 | 124 | 57 |
| 3 | Day 4 | Day 5 | 73 | 157 | 12 | 177 |
| 4 | Day 1 | Day 2 | 102 | 50 | 67 | 49 |
| 5 | Day 1 | Day 3 | 75 | 12 | 135 | 110 |
| 6 | Day 2 | Day 4 | 86 | 197 | 8 | 59 |
| 7 | Day 1 | Day 5 | 90 | 56 | 197 | 162 |
| 8 | Day 1 | Day 3 | 50 | 79 | 92 | 136 |

Table 1: **ATTac-2000**'s client preferences from game 3070. *BEV*, *SEV*, and *TEV* are *EV*s for baseball, symphony, and theater respectively.

| Client | *AD* | *DD* | Hotel | Ent'ment | Utility |
|--------|------|------|-------|----------|---------|
| 1 | Day 2 | Day 5 | LFI | B4 | 1175 |
| 2 | Day 1 | Day 2 | BGH | B1 | 1138 |
| 3 | Day 3 | Day 5 | LFI | T3, B4 | 1234 |
| 4 | Day 1 | Day 2 | BGH | None | 1102 |
| 5 | Day 1 | Day 2 | BGH | S1 | 1110 |
| 6 | Day 2 | Day 3 | BGH | B2 | 1183 |
| 7 | Day 1 | Day 5 | LFI | S2, B3, T4 | 1415 |
| 8 | Day 1 | Day 2 | BGH | T1 | 1086 |

Table 2: **ATTac-2000**'s client allocations and utilities from game 3070. Client 1's "B4" under "Ent'ment" indicates baseball on day 4.

## 3.1 Bidding Strategy

TAC was defined so as to be simple enough to have a low barrier to entry, yet complex enough to prevent tractable solution via direct game-theoretic analysis. Given that an optimal solution is not attainable, we use a principled approach that takes advantage of details of the TAC scenario. In general, **ATTac-2000** aims to be robust to the parameter space defined by TAC as well as to conceivable opponent strategies.

At every bidding opportunity, **ATTac-2000** begins by computing the most profitable allocation of goods to clients (which we shall denote $G^*$), given the goods that are currently owned and the current prices of hotels and flights. (See Section 3.3 for a caveat.) For the purposes of this computation, **ATTac-2000** allocates, but does not consider buying or selling, entertainment tickets. In most cases, $G^*$ is computed using integer linear programming, as described in Section 3.2.

**ATTac-2000**'s high-level bidding strategy is based on the following two observations:

1. While the auctions are open:

   - Obtain updated market prices.
   - Compute $G^*$: the most profitable allocation of goods given current holdings and prices.
   - Bid in 1 of 2 different modes

   **Passive:** bid to keep options open

   **Active:** at end, bid aggressively on packages

2. Allocate:

   - Compute $G^*$ with closed auctions and allocate purchased goods to clients.

Table 3: An overview of ATTAC-2000's high-level strategy.

1. Since airline prices periodically increase or decrease with equal probability, the *expected* change in price for each airline auction is $0. Indeed, it can be shown that if the airline auction is considered in isolation, waiting until the very end of the game to purchase tickets is an optimal strategy (except in the rare case that the price reaches the lowest allowed value).

2. Since hotel prices are monotonically increasing, as the game proceeds, the hotel prices approach the eventual closing prices.

Therefore, ATTAC-2000 aims to delay most of its purchases, and particularly its airline purchases, until late in the game. ATTAC-2000's high-level bidding strategy is based on the premise that it is best to delay "committing" to the current $G^*$ for as long as possible. Although it continually reevaluates $G^*$, and is therefore never technically committed to anything, the markets are such that it is rarely advantageous to change a client's travel package if it would mean wasting an airline ticket or an expensive hotel room (thus requiring additional ones to be purchased).

ATTAC-2000 accomplishes this delay of commitment by bidding in two different modes: *passive* and *active*. The *passive* mode, which lasts most of the game, is designed to keep as many options open as possible. During the passive mode, ATTAC-2000 computes the average time it takes for it to compute and place its bids, $T_b$ ($T_b$ is the average time it takes to go through one iteration of the loop in step 1 of Table 3). We found that $T_b$ ranged from 10 seconds to well over a minute, and was primarily dependent upon the server's load. Call the time left in the game $T_l$. When $T_l \leq 2 \times T_b$, ATTAC-2000 switches to its *active* mode, during which it buys the airline tickets required by the current $G^*$ and places high bids for the required hotel rooms. Note that ATTAC-2000 expects to run at most 2 bidding iterations in active mode. In fact, only 1 such iteration is necessary, but there is a huge cost to failing to complete the iteration before the end of the game. Planning for 2 active iterations leaves room for some error.

Based on the current $G^*$, its current mode, and $T_l$, ATTAC-2000 bids for flights, hotel rooms, and entertainment tickets.

### 3.1.1 FLIGHTS

While in the passive mode, ATTac-2000 does not bid in the airline auctions. In active mode, ATTac-2000 buys all currently unowned airline tickets needed for the current $G^*$. In most cases, that means that it only bids for airline tickets during its first bidding opportunity in the active mode. However, in the face of drastically changing (hotel and entertainment ticket) prices, $G^*$ could change sufficiently to necessitate purchasing additional flights, instead of simply using the ones that have already been purchased.

### 3.1.2 HOTELS

When in the passive mode, ATTac-2000 bids in the hotel auctions either to try to win hotels cheaply should the auction close early, or to try to prevent the hotel auctions from closing early. It might be advantageous to prevent a hotel auction from closing if no rooms are currently desired in order to keep open the option of switching to that hotel should future market prices warrant it.

For each hotel room of type $i$ (such as "Grand Hotel, night 3"), let $H_i$ be the number of rooms of type $i$ needed for $G^*$. Based on the current price of $i$, $P_i$, ATTac-2000 tries to acquire $n$ rooms where

$$n = \begin{cases} 8 & \text{if } P_i = 0 \text{ (only true at the outset of the game)} \\ \max(H_i, 4) & \text{if } P_i \leq 10 \\ \max(H_i, 2) & \text{if } P_i \leq 20 \\ \max(H_i, 1) & \text{if } P_i \leq 50. \end{cases}$$

If ATTac-2000's outstanding bids would already win $n$ rooms should the auction close at the current price, then ATTac-2000 does nothing: should the auction close prematurely, ATTac-2000 wins the $n$ rooms cheaply, and competitors lose the opportunity to get any rooms of type $i$ later in the game. Otherwise, ATTac-2000 bids for $n$ rooms at $1 above the current ask price. The formula for computing $n$ was selected so as to risk wasting up to $40–$50 per room type for the benefit of maintaining flexibility later in the game. The exact parameters here were chosen in an ad-hoc fashion without detailed experimentation. Our intuition is that ATTac-2000's performance is not very sensitive to their exact values.

In the active mode, ATTac-2000 bids on hotel rooms based on their marginal value within allocation $G^*$. Let $V(G^*)$ be the value of $G^*$ (the income from all clients, minus the cost of the yet-to-be-acquired goods). Let $G^{*\prime}_c$ be the optimal allocation should client $c$ fail to get its hotel rooms. Note that $G^{*\prime}_c$ might differ from $G^*$ in the distribution of entertainment tickets as well as in the flights and hotels of client $c$. ATTac-2000 bids for the hotel rooms assigned to client $c$ in $G^*$ at a price of $V(G^*) - V(G^{*\prime}_c)$. Since at this point flights are a sunk cost, this price tends to be more than $1000.

Notice that ATTac-2000 bids the full marginal utility for each hotel room required by the client's travel package. An alternative would have been to divide the marginal utility over the number of rooms in the package, which would have eliminated the risk of spending more on hotels than the itinerary is worth. On the other hand, failing to win a single hotel room is enough to invalidate the entire itinerary. ATTac-2000 bids the full marginal utility to maximize the chance that valid itineraries are obtained for all clients. In a combinatorial auction, the bidder would be able to be place a bid for the conjunction of the desired rooms and would therefore not need to choose between these two alternatives.

### 3.1.3 ENTERTAINMENT TICKETS

ATTac-2000's bidding strategy for the entertainment tickets hypothesizes that for each ticket, the opponent buy (sell) price remains constant over the course of a single game (but may vary from game to game). So as to avoid underbidding (overbidding) for that price, ATTac-2000 gradually decreases (increases) its bid over the course of the game. The initial bids are always as optimistic as possible, but by the end of the game, ATTac-2000 is willing to settle for deals that are minimally profitable. In addition, this strategy serves to hedge against ATTac-2000's early uncertainty in its final allocation of goods to clients.

On every bidding iteration, ATTac-2000 places a buy bid for each type of entertainment ticket, and a sell bid for each type of entertainment ticket that it currently owns. In all cases, the prices depend on the amount of time left in the game ($T_l$), becoming less aggressive as time goes on (see Figure 1).

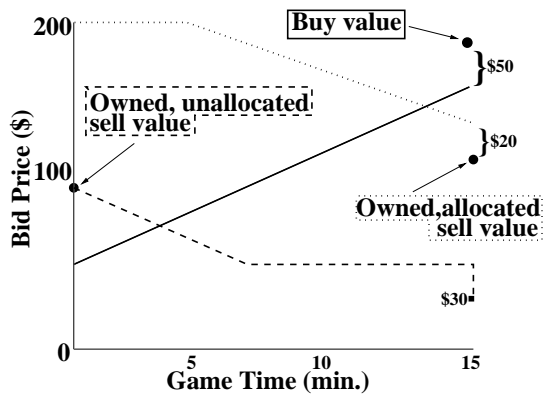

Figure 1: ATTac-2000's bidding strategy for entertainment tickets. The black circles indicate the calculated values of the tickets to ATTac-2000. The lines indicate the bid prices corresponding to those values. For example, the solid line (which increases over time) corresponds to the buy price relative to the buy value. Correspondence between the text and the lines is indicated by similar line types and boxes surrounding the text.

For each owned entertainment ticket $E$, if $E$ is assigned in $G^*$, let $V(E)$ be the value of $E$ to the client to whom it is assigned in $G^*$ ("owned, allocated sell value" in Figure 1). ATTac-2000 offers to sell $E$ for $\min(200, V(E) + \delta)$ where $\delta$ decreases linearly from 100 to 20 based on $T_l$.[1] If there is a current bid price greater than the resulting sell price, then ATTac-2000 raises its sell price to 1 cent lower than the current bid price in order to get as high a price as possible.

If $E$ is owned but not assigned in $G^*$ (because all clients are either unavailable that night or already scheduled for that type of entertainment in $G^*$), let $V(E)$ be the maximum value for $E$ over all clients, i.e. the greatest possible value of $E$ given the client profiles ("owned, unallocated sell value" in Figure 1). ATTac-2000 offers to sell $E$ for $\max(50, V(E) - \delta)$ where $\delta$ increases linearly from 0 to 50 based on $T_l$. Once again, ATTac-2000 raises its price to meet an existing bid price that is greater than its target price. This strategy reflects the increasing likelihood as the game progresses that $G^*$ will be close to the final client allocation, and thus that any currently unused tickets will not be needed in the end. When in active mode, ATTac-2000 assumes that $G^*$ is final and offers to sell any unneeded tickets for \$30 in order to obtain at least some value for them (represented by the discrete point at the bottom right in Figure 1). Below \$30, ATTac-2000 would rather waste the ticket than allow a competitor to make a large profit.

---

1. Recall that $200 is the maximum possible value of $E$ to any client under the TAC parameters.

Finally, ATTac-2000 bids to buy each type of entertainment ticket $E$ (including those that it is also offering to sell) based on the increased value that would be derived by owning $E$. Let $G^{*\prime}_E$ be the optimal allocation that would result were $E$ owned ("buy value" in Figure 1). Note that $G^{*\prime}_E$ could have different flight and hotel assignments than $G^*$ so as to make most effective use of $E$. Then, ATTac-2000 offers to buy $E$ for $V(G^{*\prime}_E) - V(G^*) - \delta$, where $\delta$ decreases linearly from 100 to 20 based on $T_l$.

All of the parameters described in this section were chosen arbitrarily without detailed experimentation. Again our intuition is that, unless opponents know and explicitly exploit these values, ATTac-2000's performance is not very sensitive to them.

## 3.2 Allocation Strategy

As is evident from Section 3.1, ATTac-2000 relies heavily on computing the current most profitable allocation of goods to clients, $G^*$. Since $G^*$ changes as prices change, ATTac-2000 needs to recompute it at every bidding opportunity. By using an integer linear programming approach, ATTac-2000 was able to compute optimal final allocations in every game instance during the tournament finals—one of only 2 entrants to do so.[2]

Most TAC participants used some form of greedy strategy for allocation (Greenwald & Stone, 2001). It is computationally feasible to quickly determine the maximum utility achievable by client 1 given a set of purchased goods, move on to client 2 with the remaining goods, etc. However, the greedy strategy can lead to suboptimal solutions. For example, consider 2 clients $A$ and $B$ with identical travel days $IAD$ and $IDD$ as well as identical entertainment values $EV$, but with $A$'s $GHV = \$50$ and $B$'s $GHV = \$150$. If the agent has exactly one of each type of hotel room for each day, the optimal assignment is clearly to assign the BGH to client $B$. However, if client $A$'s utility is optimized first, it will be assigned the BGH, leaving $B$ to stay in LFI. The agent's resulting score would be 100 less than it could have been.

As an improvement over the basic greedy strategy, we implemented a heuristic approach that implements the greedy strategy over 100 random client orderings and chooses the most profitable resulting allocation. Empirically, the resulting allocation is often optimal, and never far from optimal. In addition, it is always very quick to compute. In a set of seven games from just before the tournament, the greedy allocator was run approximately 600 times and produced allocations that averaged 99.5% of the optimal value.

---

2. As computed by Shou-de Lin of the TAC organizing team.

As the competition drew near, however, it became clear that every point would count. We therefore implemented an allocation strategy that is guaranteed to find the optimal allocation of goods.[3] The integer linear programming approach used by **ATTac-2000** works by defining a set of variables, constraints on these variables, and an objective function. An assignment to the variables represents an allocation to the clients and the constraints ensure that the allocation is legal. The objective function encodes the fact that we seek the allocation with maximum value (utility minus cost).

The following notation is needed to describe the integer linear program. The formal notation is included for completeness; an equivalent English description follows each equation. The symbol $c$ is a client (1 through 8). The symbol $f$ is a feasible travel package, which consists of: the arrival day $AD(f)$ (1 through 4); the departure day $DD(f)$ (2 through 5), and the choice of hotel $H(f)$ (BGH or LFI). There are 20 such travel packages. Symbol $e$ is an entertainment ticket, which consists of: the day of the event $D(e)$ (1 through 4), and the type of the event $T(e)$ (baseball $b$, symphony $s$, or theater $t$). There are 12 different entertainment tickets. Symbol $r$ is a resource ($AD$, $DD$, BGH, or LFI).

Using this notation, the 272 variables are: $P(c, f)$, which indicates whether client $c$ will be allocated feasible travel package $f$ (160 variables); $E(c, e)$, which indicates whether client $c$ will be allocated entertainment ticket $e$ (96 variables); and, $B_r(d)$ is the number of copies of resource $r$ we would like to buy for day $d$ (16 variables).

There are also several constants that define the problem: $o_r(d)$ is the number of tickets of resource $r$ currently owned for day $d$, $p_r(d)$ is the current price for resource $r$ on day $d$, $u_P(c, f)$ is utility to customer $c$ for travel package $f$, and $u_E(c, e)$ is the utility to customer $c$ for entertainment ticket $e$.

Given this notation, the objective is to maximize utility minus cost

$$\sum_{c,f} u_P(c, f)P(c, f) + \sum_{c,e} u_E(c, e)E(c, e)$$

$$- \sum_{d \in \{2,3,4,5\}} p_{DD}(d)B_{DD}(d)$$

$$- \sum_{d \in \{1,2,3,4\}, r \in \{\text{BGH,LFI},AD\}} p_r(d)B_r(d)$$

subject to the following 188 constraints:

- For all $c$, $\sum_f P(c, f) \leq 1$: No client gets more than one travel package (8 constraints).

- For all $d \in \{1, 2, 3, 4\}$,

$$\sum_c \sum_{f|AD(f)=d} P(c, f) \leq o_{AD}(d) + B_{AD}(d),$$

For all $d \in \{1, 2, 3, 4\}$ and $h \in \{\text{BGH, LFI}\}$,

$$\sum_c \sum_{f|H(f)=h \ \& \ AD(f) \leq d < DD(f)} P(c, f) \leq o_h(d) + B_h(d),$$

---

3. The general allocation problem is NP-complete, as it is equivalent to the set-packing problem (Garey & Johnson, 1979). Exhaustive search is computationally intractable even with only 8 clients.

For all $d \in \{2, 3, 4, 5\}$,

$$\sum_c \sum_{f | DD(f) = d} P(c, f) \leq o_{DD}(d) + B_{DD}(d):$$

The demand for resources from the selected travel packages must not exceed the sum of the owned and bought resources (16 constraints).

- For all $e$, $\sum_c E(c, e) \leq o_E(e)$: The total quantity of each entertainment ticket allocated does not exceed what is owned (12 constraints).

- For all $c$ and $e$, $\sum_{f | AD(f) \leq D(e) < DD(f)} P(c, f) \geq E(c, e)$: An entertainment ticket can only be used if its day is between the arrival and departure day of the selected travel package (96 constraints).

- For all $c$ and $d \in \{1, 2, 3, 4\}$, $\sum_{e | D(e) = d} E(c, e) \leq 1$: Each client can only use one entertainment ticket per day (32 constraints).

- For all $c$ and $y \in \{b, s, t\}$, $\sum_{e | T(e) = y} E(c, e) \leq 1$: Each client can only use each type of entertainment ticket once (24 constraints).

- All variables are integers.

The solution to the resulting integer linear program is a value-maximizing allocation of owned resources to customers along with a list of resources that need to be purchased. Using the linear programming package "LPsolve", ATTac-2000 is usually able to find the globally optimal solution in under one second on a 650 MHz Pentium II.

Note that this is not by any means the only possible formulation of the allocation. Greenwald, Boyan, Kirby, and Reiter (2001) studied a variant and found that it performed extremely well on a collection of large, random allocation problems.

The above approach is guaranteed to find the optimal allocation, and usually does so quickly. However, since integer linear programming is an NP-complete problem, some inputs can lead to significantly longer solution times. In a sample of 32 games taken shortly before the finals, the allocator was called 1866 times. In 93% of the cases, the optimization took a second or less. Less than 1% took 6 or more seconds. However, the 3 longest running times were all over a minute and all came from the same game. ATTac-2000 used the strategy that if an integer linear program takes 6 or more seconds to solve, the above-mentioned greedy strategy over random client orderings is used as a fall-back strategy for the rest of the game. This fall-back strategy was not needed during the tournament finals.

### 3.3 Adaptivity

In a TAC game instance, the only information available to agents is the ask prices—individual bids are not visible. After each game, transaction-by-transaction data is available, but the lack of within-game information precluded competitors from using detailed models of opponent strategies in decision making. ATTac-2000 instead adapts its behavior on-line in three different ways: adaptable timing of bidding modes; adaptable allocation strategy; and adaptable hotel bidding.

### 3.3.1 Timing of Bidding Modes

ATTac-2000 decides when to switch from the passive to the active bidding mode based on the observed server latency $T_b$ during the current game instance (see Section 3.1).

### 3.3.2 Allocation

ATTac-2000 adapts its allocation strategy based on the amount of time it takes for the integer linear programming approach to determine optimal allocations in the current game instance (see Section 3.2).

### 3.3.3 Hotel Bidding

Perhaps most significantly, ATTac-2000 predicts the closing prices of hotel auctions based on their closing prices in previous games. Hotel bidding in TAC was particularly challenging due to the extreme volatility of prices near the end of the game. As stated in Section 3.1.2, at the end of the game ATTac-2000 bids the marginal utility for each desired hotel room, which was often in excess of $1000.

During the preliminary competition, few agents bid their marginal utilities on hotel rooms. Those that did, however, generally dominated their competitors; such agents were *high-bidders*, bidding ∼ $1000, always winning the hotels on which they bid, but paying far less than their bids. Having observed a dominant strategy during the preliminary rounds, most agents, including ATTac-2000, adopted this high-bidding strategy during the actual competition. The result was many negative scores, as prices skyrocketed in the last moments of the game once there were 16 high bids for a given room.

In Section 3.1, we stated that ATTac-2000 computes $G^*$ based on the current prices of the hotel rooms. Should the prices eventually become very high, ATTac-2000 would either end up paying too high a price for the hotel rooms or else fail to get travel packages for some of its clients. The only alternative was to avoid counting on obtaining contentious hotel rooms.

Since strategies were changing up to the last minute before the finals, there was no way to identify a priori which hotels would be most contentious or whether hotel prices would actually skyrocket in the tournament. Therefore, ATTac-2000 divided the 8 hotel rooms into 4 equivalence classes, exploiting symmetries in the game (hotel rooms on days 1 and 4 should be equally in demand as should rooms on days 2 and 3), assigned priors to the expected closing prices of these rooms, and then adjusted these priors based on the observed closing prices during the tournament.

As expected, the Grand Hotel on days 2 and 3 turned out to be most contentious during the finals. Le Fleabag Inn on the same days was also fairly contentious. Whenever the actual price for a hotel was less than the predicted closing price, ATTac-2000 used the predicted hotel closing price for computing all of its allocation values.

One additional method for predicting whether hotel prices would skyrocket in a given game is to notice who the participants are and whether or not they tended to be high-bidders in past games (see Figure 2). Although such information was not available via the server's API, a game's participants were always published beforehand on the TAC web page. By automatically downloading this information from the web (a practice whose ethicality was questioned at the competition), and matching against a precompiled database of which agents were high-bidders in the past, ATTac-2000 would only use the predicted hotel closing prices in games with 3 or more high-bidders involved: in games with fewer high-bidders, the prices of hotel rooms almost never skyrocketed[4]. As it turned out, all but one of ATTac-2000's games in the semi-finals, and all games in the finals, involved several high-bidders, thus triggering the use of predicted hotel closing prices.

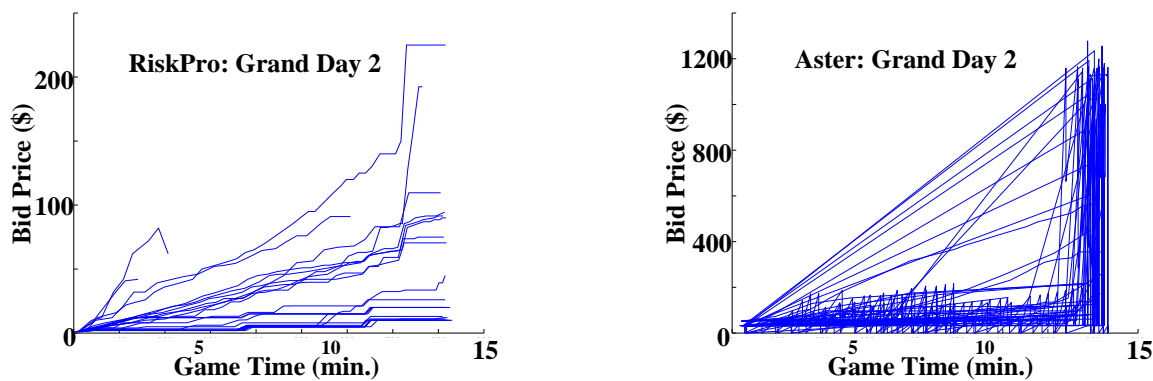

Figure 2: Graphs of two different agents' bidding patterns over many games. Each line represents one game's worth of bidding in a single auction. Left: RiskPro never bids over $250 in the games plotted. Right: Aster consistently bids over $1000 for rooms.

Empirical testing (Section 4) indicates that this strategy is extremely beneficial in situations in which hotel prices do indeed escalate, while it does not lead to significantly degraded performance when they do not.

## 4. Results

TAC consisted of a preliminary round that ran over the course of a week and involved roughly 80 games for each of the 22 participants. The top 12 finishers were invited to the semi-finals and finals in Boston, MA on July 8th. Since agents and conditions were constantly changing, and since only 13 games were played by each agent in the semi-finals and finals, the competition does not provide a controlled testing environment. In this section, we describe ATTac-2000's success in the tournament, but also present empirical results of controlled tests that demonstrate the effectiveness and robustness of ATTac-2000's adaptive strategy.

### 4.1 The Competition

ATTac-2000's scores in the 88 preliminary-round games ranged from −3000 to over 4500 (mean 2700, std. dev. 1600). A good score in a game instance is in the 3000 to 4000 range. We noticed that there were many very bad scores (12 less than 1000 and seven less than 0).

---

4. With just 2 high-bidders, the only way to have the price escalate would be if they bid for a combined total of 16 rooms of the same hotel type. That could only happen if all of their clients were to stay in the same hotel on the same night, a very unlikely scenario given the TAC parameters.

This is largely the result of ATTac-2000 not yet being imbued with its adaptive timing of bidding modes. During the preliminary round, ATTac-2000 shifted from passive to active bidding mode with 50 seconds left in the game instance. While 50 seconds is usually plenty of time to allow for at least 2 iterations through ATTac-2000's bidding loop, there were occasions in which the network and server lags were such that it would take more than 50 seconds to obtain updated market prices and submit bids. In this case, ATTac-2000 would either fail to buy airline tickets, or worse still, would buy airline tickets but not get the final hotel bids in on time. Noticing that the server lag tended to be consistent within a game instance (perhaps due to the traffic patterns generated by the participating agents), we introduced the adaptive timing of bidding modes described in Section 3.3. After this change, ATTac-2000 was always able to complete at least one, and usually two, bidding loops in the active bidding phase.

The adaptive allocation strategy never came into play in the finals, as ATTac-2000 was able to optimally solve all of the allocation problems that came up during the finals very quickly using the integer linear programming method.

However, the adaptive hotel bidding did play a big role. ATTac-2000 performed as well as the other best teams in the early TAC games when hotel prices (surprisingly) stayed low, and then out-performed the competitors in the final games of the tournament when hotel prices suddenly rose to high levels. Indeed, in the last 2 games, some of the popular hotels closed at over $400. ATTac-2000 steered clear of these hotel rooms more effectively than its closest competitors.

Table 4 shows the scores of the 8 TAC finalists (Wellman et al., 2001). ATTac-2000's consistency (std. dev. 443 as opposed to 1600 in the preliminaries) is apparent: it avoided having any disastrous games, presumably due in large part to its adaptivity regarding timing and hotel bidding.

| Rank | Team | Avg. Score | Std. Dev. | Institution |
|------|------|-----------|-----------|-------------|
| 1 | ATTac-2000 | 3398 | 443 | AT&T Labs − Research |
| 2 | RoxyBot | 3283 | 545 | Brown University, NASA Ames Research |
| 3 | aster | 3068 | 493 | STAR Lab, InterTrust Technologies |
| 4 | umbctac1 | 3051 | 1123 | University of Maryland at Baltimore County |
| 5 | ALTA | 2198 | 1328 | Artificial Life, Inc. |
| 6 | m_rajatish | 1873 | 1657 | University of Tulsa |
| 7 | RiskPro | 1570 | 1607 | Royal Inst. Technology, Stockholm University |
| 8 | T1 | 1167 | 1593 | Swedish Inst. Computer Science, Industilogik |

Table 4: The scores of the 8 TAC finalists in the semi-finals and finals (13 games).

## 4.2 Controlled Testing

In order to evaluate ATTac-2000's adaptive hotel bidding strategy in a controlled manner, we ran several game instances with ATTac-2000 playing against two variants of itself:

1. *High-bidder* always computed $G^*$ based on the current hotel prices (as opposed to using priors and averages of past closing prices).

2. *Low-bidder* always computed $G^*$ as in variant 1, but also only bid for hotel rooms at $50 over the current ask price (as opposed to the marginal utility, which tended to be more than $1000).

At the extremes, with ATTac-2000 and 7 *high-bidders* playing, at least one hotel price skyrockets in every game since all agents bid very high for the hotel rooms. On the other hand, with ATTac-2000 and 7 *low-bidders* playing, hotel prices never skyrocket since all agents but ATTac-2000 bid close to the ask price. Our goal was to measure whether ATTac-2000 could perform well in both extreme scenarios as well as various intermediate ones. Table 5 summarizes our results.

| #high | agent 2 | agent 3 | agent 4 | agent 5 | agent 6 | agent 7 | agent 8 |
|-------|---------|---------|---------|---------|---------|---------|---------|
| 7 (14) | ←— | 9526 | ———————————————————————→ | | | | |
| 6 (87) | ←— | 10679 | ——————————————————————→ | | | | 1389 |
| 5 (84) | ←— | 10310 | ——————————————→ | | ←— | | 2650 |
| 4 (48) | ←— | 10005 | ————————→ | | ←———————— | | 4015 |
| 3 (21) | ←— | 5067 | —→ | ←———————————— | | | 3639 |
| 2 (282) | ←— | *209* | ←—————————————————— | | | | 2710 |

Table 5: The difference between ATTac-2000's score and the score of each of the other seven agents averaged over all games in a controlled experiment. All differences are statistically significant at the 0.001 level, except the one marked in italics. Each row corresponds to a different number of high-bidders (excluding ATTac-2000 itself). The first column presents the number of high-bidders as well as the number of experiments we ran for that scenario (in parentheses). The column labeled "agent $i$" shows how much better ATTac-2000 did on average than agent $i$. Scores above the stair-step line are for high-bidders (variant 1) and scores below the line are for low-bidders (variant 2). Results for identical agents are averaged to obtain a single average score difference for each type of agent in each row. In all cases, ATTac-2000 beats the other agents.

Each row of Table 5 corresponds to a different number of high-bidders in the game; for example, the row labeled with 4 high-bidders corresponds to ATTac-2000 playing with 4 copies of variant 1 and 3 copies of variant 2. Results for identical agents are averaged to obtain a single average score difference for each type of agent in each row. In the first column, we also show in parentheses the number of games played for the results in each row—each row reflects a different number of runs. In all cases, we ran enough game instances to achieve statistically significant results. However, in some cases we ran more instances than turned out to be required. The column labeled agent $i$ shows the difference between ATTac-2000's score and the score of agent $i$ averaged over all games. In all scenarios, these differences are positive, showing that ATTac-2000 outscored all other agents on average.[5] Statistical significance was computed from paired T-tests; all results are significant at the 0.001 level except for the one marked in italics. As mentioned before, if the number of high-bidders is greater than or equal to 3, we expect the price for contentious hotels to rise, and in all such scenarios ATTac-2000 significantly outperforms all the other agents. The large score differences appearing in the top rows of Table 5 are mainly due to the fact that the other agents get large, negative scores since they end up buying many expensive hotel rooms.

In these experiments, ATTac-2000 always uses its adaptive hotel price expectations, even when there are only 2 high-bidders. In the last row, when the number of high-bidders is 2, very little bidding up of hotel prices is expected and in this case, we do not get statistical significance relative to the two high-bidders (agent 2 and agent 3), since their strategies are nearly identical to ATTac-2000's in this case. We do get high statistical significance relative to all the other agents (copies of variant 2), however. Thus, ATTac-2000's adaptivity to hotel prices seems to help a lot when hotel prices do skyrocket and does not seem to prevent ATTac-2000 from winning on average when they don't.

The results of Table 5 provide strong evidence for ATTac-2000's ability to adapt robustly to varying number of competing agents that bid up hotel prices near the end of the game. Note that ATTac-2000 is *not* designed to perform well against itself. If 8 copies of ATTac-2000 play against each other repeatedly, they will all favor the same hotel rooms and thus consistently *all* get large negative scores. It would be interesting to determine whether there exists a strategy that is both harmful to ATTac and beneficial to the adversary.

## 5. Related Work

Although there has been a good deal of research on auction theory, especially from the perspective of auction mechanisms (Klemperer, 1999), studies of autonomous bidding agents and their interactions are relatively few and recent. TAC is one example. FM97.6 is another auction test-bed, which is based on fishmarket auctions (Rodriguez-Aguilar, Martin, Noriega, Garcia, & Sierra, 2001). Automatic bidding agents have also been created in this domain (Gimenez-Funes, Godo, Rodriguez-Aguiolar, & Garcia-Calves, 1998). There have been a number of studies of agents bidding for a single good in multiple auctions (Ito, Fukuta, Shintani, & Sycara, 2000; Anthony, Hall, Dang, & Jennings, ; Preist, Bartolini, & Phillips, 2001) Outside of, but related to, the auction scenario, automatic shopping and pricing agents for internet commerce have been studied within a simplified model (Greenwald & Kephart, 1999).

Twenty-two agents from 6 countries entered TAC, 12 of which qualified to compete in the semi-finals and finals in Boston. The designs of these agents were motivated by a wide variety of research interests including machine learning, artificial life, experimental economics, real-time systems, and choice theory (Greenwald & Stone, 2001).

Our own approach was motivated by our research interests in multiagent learning (Littman, 1994; Stone, 2000; Singh, Kearns, & Mansour, 2000). Based on the problem description, we expected to find several learning opportunities in the domain. As noted above, detailed opponent modeling was precluded by the system dynamics. Nonetheless, ATTac-2000's adaptivity is one of the keys to its success, particularly in avoiding skyrocketing hotels.

---

5. In general, ATTac-2000's average score decreased with increasing numbers of high-bidders, as games became more volatile.

The 2nd and 3rd place agents both used a different strategy to prepare for the possibility of skyrocketing hotels. Rather than avoiding popular hotels entirely by tracking closing prices across game instances, they both discouraged their agents from bidding for too many of any particular hotel room, thus spreading their demand across the rooms (Greenwald & Stone, 2001). While such a strategy is safer in the limit (i.e., it continues to work even if everyone uses it), it has a greater potential to cost the agent in the event that hotel prices do not skyrocket, since the agent will still distribute its demand to the less desirable rooms. On the other hand, ATTac-2000 would notice that the prices are not skyrocketing and thus bid for the optimal travel packages given current prices.

## 6. Conclusion and Future Work

TAC-2000 was the first autonomous bidding agent competition. While it was a very successful event, some minor improvements would increase its interest from a multiagent learning perspective.

- Currently, there is no incentive to buy airline tickets until the end of the game. Were the price of flights to tend to increase, or were supply limited, agents would have to balance the advantage of keeping their options open against the savings of committing to travel packages earlier[6].

- The information structure of the TAC setup was such that it was impossible to observe the bidding patterns of individual agents during games. Nonetheless, the strategic behavior of individual agents often profoundly affected market dynamics—particularly in the hotel auctions. It seems that it would be beneficial to be able to directly observe the behavior of each individual agent. Were there to be information available regarding the bidding behavior of the agents during the game (such that other agents could infer clients' preferences, and therefore market supply, demand, and prices), TAC agents would potentially be able to learn to predict market behavior as a game proceeds.

With or without these modifications, we hope to be able to participate in future TACs, with the goal of adding additional adaptive elements to ATTac-2000.

Another direction of future research is to apply the lessons learned from TAC to real simultaneous interacting auctions. It is straightforward to write bidding agents to participate in on-line auctions for a single good if the value to the client is fixed ahead of time: the agent can bid slightly over the ask price until the auction closes or the price exceeds the value. However, when the values of multiple goods interact, such as is the case in TAC, agent deployment is not nearly so straightforward.

One such real application is the Federal Communications Commission's auctioning off of radio spectrum (Weber, 1997; Cramton, 1997). Especially for companies that are trying to achieve national coverage, the values of the different licenses interact in complex ways. Perhaps autonomous bidding agents will be able to affect bidding strategies in such future auctions. Indeed, in related research we have begun down this path by creating straight-forward bidding agents in a realistic FCC Auction Simulator (Csirik, Littman, Singh, & Stone, 2001).

---

6. This change has been adopted in the specification of TAC-01.

In a more obvious application, an extended version of ATTac-2000 could potentially become useful to real travel agents, or to end users who wish to create their own travel packages.

## Acknowledgements

We would like to thank the TAC team at the University of Michigan, including Michael Wellman, Peter Wurman, Kevin O'Malley, Daniel Reeves, and William Walsh, for constructing the TAC server and responding promptly and cordially to our many requests while conducting the research reported here. We would also thank the anonymous reviewers for their helpful comments and suggestions.